\documentclass[11pt,a4paper]{article}
\usepackage[hyperref]{acl2019}
\usepackage{times}
\usepackage{latexsym}

\usepackage{url}

\usepackage{todonotes}
\usepackage{linguex}
\usepackage{todonotes}
\usepackage{enumitem}
\usepackage{booktabs}
\usepackage{colortbl}
\usepackage{rotating}
\usepackage{aswhitedefs}

\aclfinalcopy 
\def\aclpaperid{***} 

\title{Fine-Grained Temporal Relation Extraction}

\author{Siddharth Vashishtha  \\
  University of Rochester  \\
  \\\And
  Benjamin Van Durme \\
  Johns Hopkins University \\
   \\\And
  Aaron Steven White \\
  University of Rochester 
}

\date{}

\begin{document}
\maketitle
\setlength{\Exlabelsep}{0em}
\setlength{\Extopsep}{.2\baselineskip}

\setlength{\SubExleftmargin}{1.3em}

\begin{abstract}
We present a novel semantic framework for modeling temporal relations and event durations that maps pairs of events to real-valued scales. We use this framework to construct the largest temporal relations dataset to date, covering the entirety of the Universal Dependencies English Web Treebank.  We use this dataset to train models for jointly predicting fine-grained temporal relations and event durations. We report strong results on our data and show the efficacy of a transfer-learning approach for predicting categorical relations.

\end{abstract}

\section{Introduction}
Natural languages provide a myriad of formal and lexical devices for conveying the temporal structure of complex events---e.g. tense, aspect, auxiliaries, adverbials, coordinators, subordinators, etc. Yet, these devices are generally insufficient for determining the fine-grained temporal structure of such events. Consider the narrative in \ref{ex:intronarrative}.

\ex. At 3pm, a boy \textbf{broke} his neighbor's window. He was \textbf{running away}, when the neighbor \textbf{rushed out} to \textbf{confront} him. His parents were \textbf{called} but couldn't \textbf{arrive} for two hours because they \textbf{were still at work}. \label{ex:intronarrative}

\begin{figure}[t]
\includegraphics[width=\columnwidth]{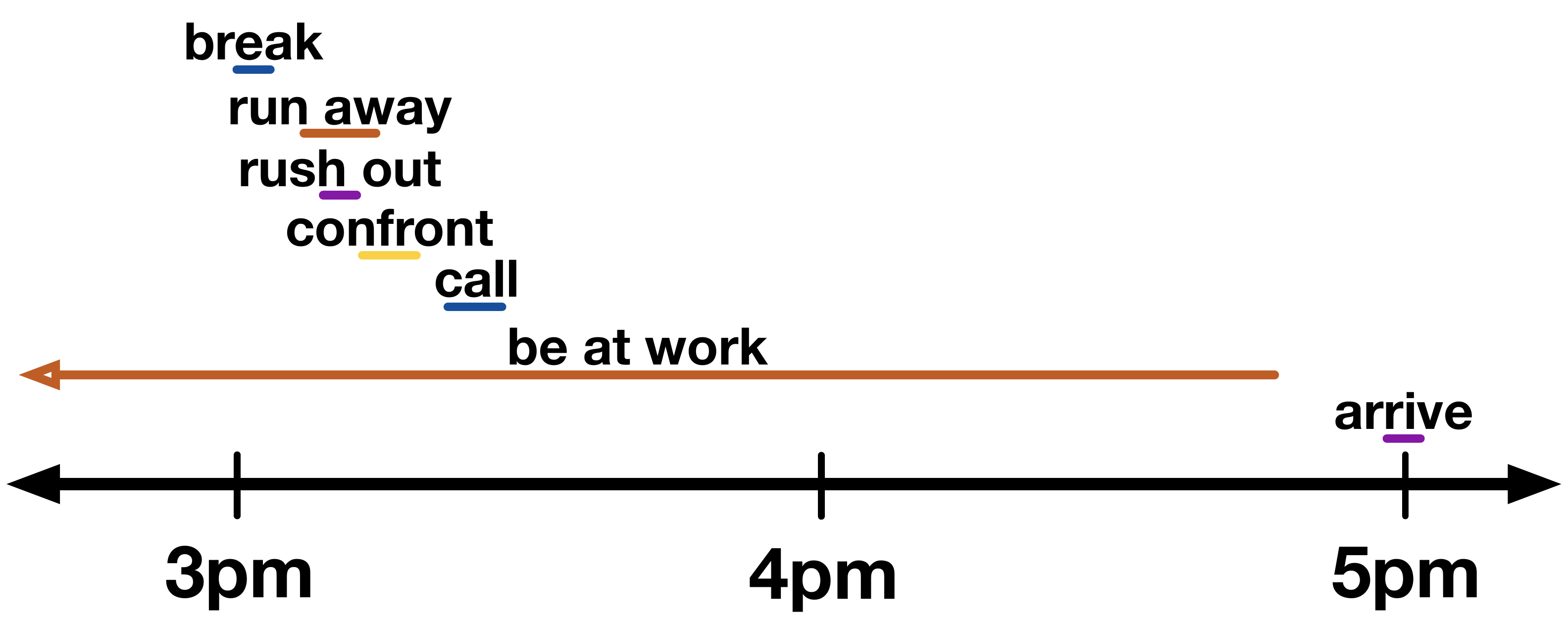}
\vspace{-6mm}
\caption{A typical timeline for the narrative in \ref{ex:intronarrative}.}
\label{fig:introtimeline}
\vspace{-6mm}
\end{figure}

Most native English speakers would have little difficulty drawing a timeline for these events, likely producing something like that in Figure \ref{fig:introtimeline}. But how do we know that the breaking, the running away, the confrontation, and the calling were short, while the parents being at work was not? And why should the first four be in sequence, with the last containing the others? 

The answers to these questions likely involve a complex interplay between linguistic information, on the one hand, and common sense knowledge about events and their relationships, on the other ~\citep{minsky1975framework, schank1975scripts, lamport1978time, allen1985common, hobbs1987commonsense, hwang1994interpreting}. But it remains a question how best to capture this interaction.

A promising line of attack lies in the task of temporal relation extraction. Prior work in this domain has approached this task as a classification problem, labeling pairs of event-referring expressions---e.g. \textit{broke} or \textit{be at work} in \ref{ex:intronarrative}---and time-referring expressions---e.g. \textit{3pm} or \textit{two hours}---with categorical temporal relations~\citep{pustejovsky2003timebank, styler2014temporal, minard2016meantime}. The downside of this approach is that time-referring expressions must be relied upon to express duration information. But as \ref{ex:intronarrative} highlights, nearly all temporal duration information can be left implicit without hindering comprehension, meaning these approaches only explicitly encode duration information when that information is linguistically realized.

In this paper, we develop a novel framework for temporal relation representation that puts event duration front and center. Like standard approaches using the TimeML standard, we draw inspiration from \citeauthor{allen1983maintaining}'s (\citeyear{allen1983maintaining}) seminal work on interval representations of time. But instead of annotating text for categorical temporal relations, we map events to their likely durations and event pairs directly to real-valued relative timelines. This change not only supports the goal of giving a more central role to event duration, it also allows us to better reason about the temporal structure of complex events as described by entire documents.

We first discuss prior work on temporal relation extraction (\S\ref{sec:background}) and then present our framework and data collection methodology (\S\ref{sec:datacollection}). The resulting dataset---Universal Decompositional Semantics Time (UDS-T)---is the largest temporal relation dataset to date, covering all of the Universal Dependencies \cite{silveira2014gold, de_marneffe_universal_2014,nivre_universal_2015} English Web Treebank \citep{bies2012english}. We use this dataset to train a variety of neural models (\S\ref{sec:model}) to jointly predict event durations and fine-grained (real-valued) temporal relations (\S\ref{sec:experiments}), yielding not only strong results on our dataset, but also competitive performance on TimeML-based datasets (\S\ref{sec:results}).\footnote{Data and code are available at \url{http://decomp.io/}.}

\section{Background}
\label{sec:background}
We review prior work on temporal relations frameworks and temporal relation extraction systems.

\paragraph{Corpora} 

Most large temporal relation datasets use the TimeML standard~\citep{pustejovsky2003timebank, styler2014temporal, minard2016meantime}. TimeBank is one of the earliest large corpora built using this standard, aimed at capturing `salient' temporal relations between events~\citep{pustejovsky2003timebank}. The TempEval competitions build on TimeBank by covering relations between all the events and times in a sentence. 

Inter-sentential relations, which are necessary for document-level reasoning, have not been a focus of the TempEval tasks, though at least one sub-task does address them~\citep[][and see \citealt{chambers2014dense}]{verhagen2007semeval, verhagen2010semeval, uzzaman2013semeval}. Part of this likely has to do with the sparsity inherent in the TempEval event-graphs. This sparsity has been addressed with corpora such as the TimeBank-Dense, where annotators label all local-edges irrespective of ambiguity~\citep{cassidy2014annotation}. TimeBank-Dense does not capture the complete graph over event and time relations, instead attempting to achieve completeness by capturing all relations both within a sentence and between neighboring sentences. We take inspiration from this work for our own framework.

This line of work has been further improved on by frameworks such as Richer Event Description (RED), which uses a multi-stage annotation pipeline where various event-event phenomena, including temporal relations and sub-event relations are annotated together in the same datasets~\citep{o2016richer}. Similarly, \citet{hong2016building} build a cross-document event corpus which covers fine-grained event-event relations and roles with more number of event types and sub-types~\citep[see also][]{fokkens2013gaf}. 

\paragraph{Models}

Early systems for temporal relation extraction use hand-tagged features modeled with multinomial logistic regression and support vector machines~\citep{mani2006machine, bethard2013cleartk, lin2015multilayered}. Other approaches use combined rule-based and learning-based approaches~\citep{d2013classifying} and sieve-based architectures---e.g. CAEVO~\citep{chambers2014dense} and CATENA~\citep{mirza2016catena}. Recently, \citet{ning2017structured} use a structured learning approach and show significant improvements on both TempEval-3 \citep{uzzaman2013semeval} and TimeBank-Dense \citep{cassidy2014annotation}. \citet{ning2018joint} show further improvements on TimeBank-Dense by jointly modeling causal and temporal relations using Constrained Conditional Models and formulating the problem as an Interger Linear Programming problem.

Neural network-based approaches have used both recurrent~\citep{tourille2017neural, cheng2017classifying, leeuwenberg2018temporal} and convolutional architectures~\citep{dligach2017neural}. Such models have furthermore been used to construct document timelines from a set of predicted temporal relations \citep{leeuwenberg2018temporal}. Such use of pairwise annotations can result in inconsistent temporal graphs, and efforts have been made to avert this issue by employing temporal reasoning~\citep{chambers2008jointly, yoshikawa2009jointly, denis2011predicting, do2012joint, laokulrat2016stacking, ning2017structured, leeuwenberg2017structured}. 

Other work has aimed at modeling event durations from text~\citep{pan2007modeling, gusev2011using, williams2012extracting}, though this work does not tie duration to temporal relations \citep[see also][]{filatova2001assigning}. Our approach combines duration and temporal relation information within a unified framework, discussed below.

\section{Data Collection}
\label{sec:datacollection}
We collect the Universal Decompositional Semantics Time (UDS-T) dataset, which is annotated on top of the Universal Dependencies \cite{silveira2014gold, de_marneffe_universal_2014,nivre_universal_2015} English Web Treebank \citep{bies2012english} (UD-EWT). The main advantages of UD-EWT over other similar corpora are: (i) it covers text from a variety of genres; (ii) it contains gold standard Universal Dependency parses; and (iii) it is compatible with various other semantic annotations which use the same predicate extraction standard \cite{white_universal_2016,zhang2017evaluation,rudinger2018neural,govindarajan_gener_2019}. Table \ref{tab:datasetsize} compares the size of UDS-T against other temporal relations datasets. 

\begin{table}[t]
\small
\begin{center}
\begin{tabular}%
  {>{\raggedright\arraybackslash}p{2.5cm}%
   >{\raggedleft\arraybackslash}p{1.5cm}%
   >{\raggedleft\arraybackslash}p{2cm}%
  }
\toprule
Dataset & \#Events & \#Event-Event Relations\\
\midrule
TimeBank & 7,935 & 3,481\\
TempEval 2010 & 5,688 & 3,308 \\
TempEval 2013 & 11,145 & 5,272\\
TimeBank-Dense & 1,729 & 8,130\\
\citet{hong2016building} & 863 & 25,610\\
\textbf{UDS-T} & \textbf{32,302} & \textbf{70,368} \\
\bottomrule
\end{tabular}
\end{center}
\vspace{-3mm}
\caption{Number of total events, and event-event temporal relations captured in various corpora}
\label{tab:datasetsize}
\vspace{-5mm}
\end{table}

\paragraph{Protocol design}

Annotators are given two contiguous sentences from a document with two highlighted event-referring expressions (predicates). They are then asked (i) to provide relative timelines on a bounded scale for the pair of events referred to by the highlighted predicates; and (ii) to give the likely duration of the event referred to by the predicate from the following list: \textit{instantaneous, seconds, minutes, hours, days, weeks, months, years, decades, centuries, forever}. In addition, annotators were asked to give a confidence ratings for their relation annotation and each of their two duration annotation on the same five-point scale - \textit{not at all confident} (0), \textit{not very confident} (1), \textit{somewhat confident} (2), \textit{very confident} (3), \textit{totally confident} (4). 

\begin{figure}[t]
\includegraphics[width=\columnwidth, scale=1]{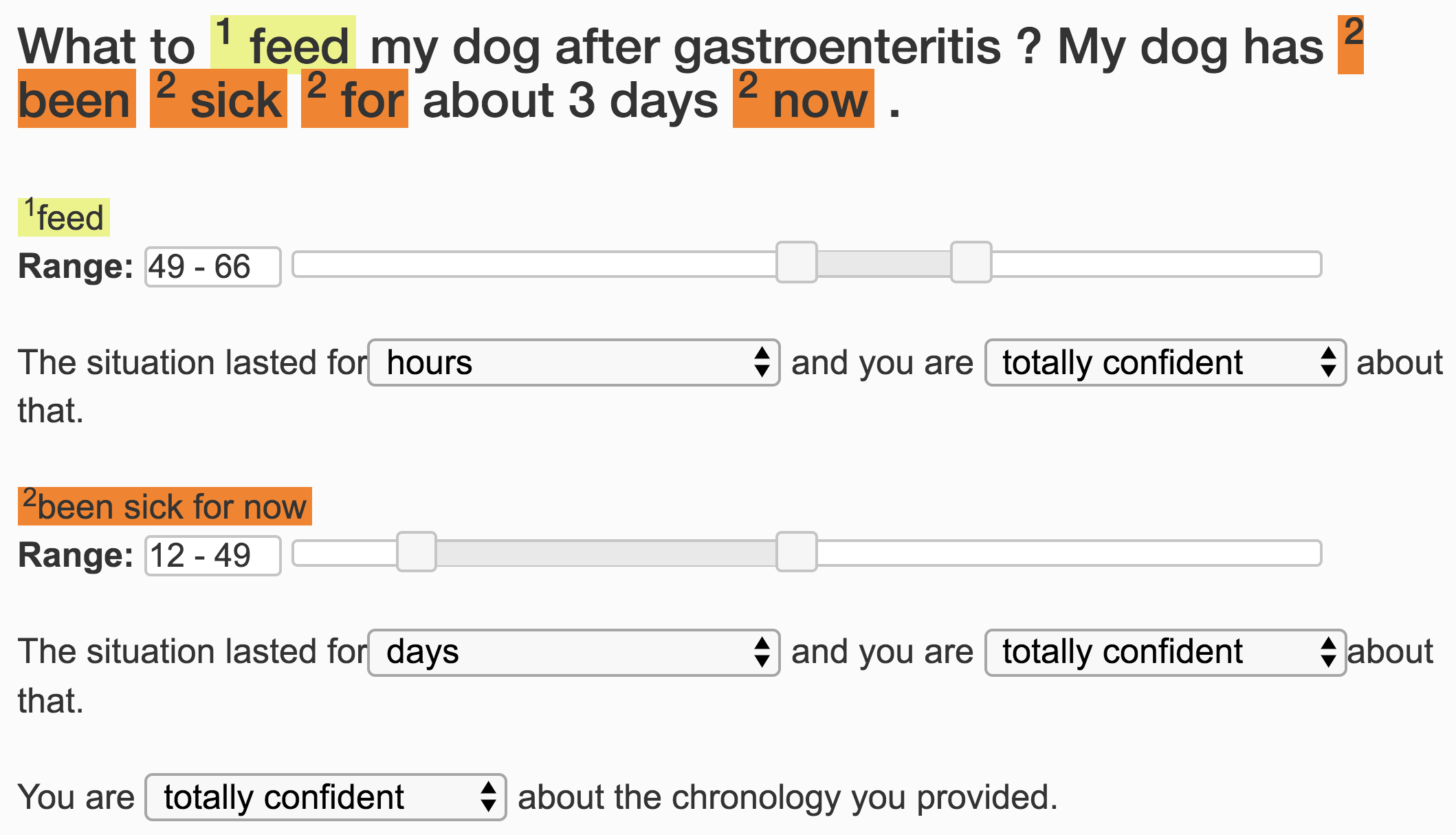}
\vspace{-7mm}
\caption{An annotated example from our protocol}
\label{fig:protocolexample}
\vspace{-6mm}
\end{figure}

An example of the annotation instrument is shown in Figure \ref{fig:protocolexample}.  Henceforth, we refer to the situation referred to by the predicate that comes first in linear order (\textit{feed} in Figure \ref{fig:protocolexample}) as $e_1$ and the situation referred to by the predicate that comes second in linear order (\textit{sick} in Figure \ref{fig:protocolexample}) as $e_2$.

\paragraph{Annotators}

We recruited 765 annotators from Amazon Mechanical Turk to annotate predicate pairs in groups of five. Each predicate pair contained in the UD-EWT train set was annotated by a single annotator, and each in the UD-EWT development and test sets was annotated by three.

\paragraph{Predicate extraction} 

We extract predicates from UD-EWT using PredPatt \cite{white_universal_2016,zhang2017evaluation}, which identifies 33,935 predicates from 16,622 sentences. 
We concatenate all pairs of adjacent sentences in the documents contained in UD-EWT, allowing us to capture inter-sentential temporal relations. Considering all possible pairs of predicates in adjacent sentences is infeasible, so we use a heuristic to capture the most interesting pairs. (See Appendix \ref{sec:data_collect_appendix} for details.)

\paragraph{Normalization}
\begin{figure}[htb]
\includegraphics[width=\columnwidth, scale=1]{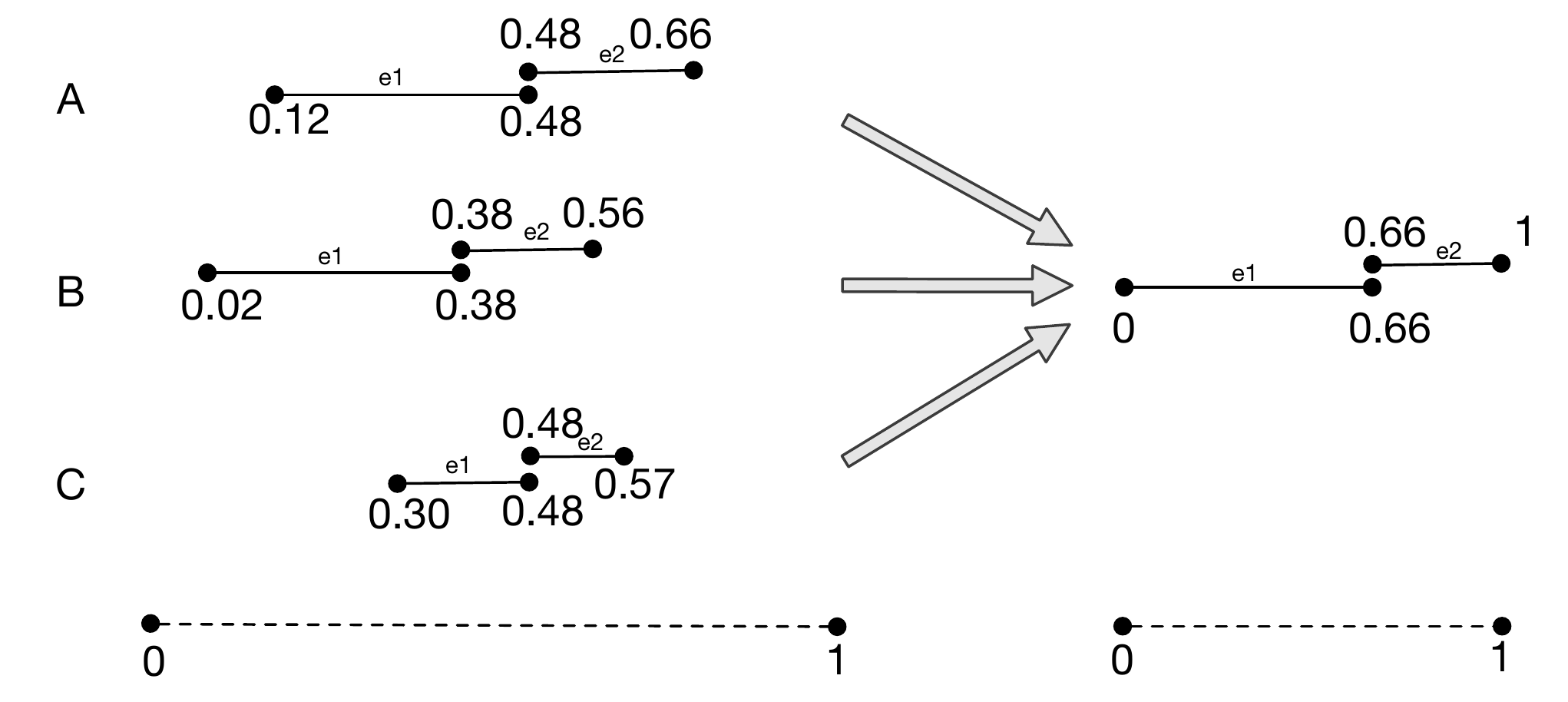}
\vspace{-7mm}
\caption{Normalization of slider values}
\label{fig:normalization}
\vspace{-3mm}
\end{figure}
We normalize the slider responses for each event pair by subtracting the minimum slider value from all values, then dividing all such shifted values by the maximum value (after shifting). This ensures that the earliest beginning point for every event pair lies at 0 and that the right-most end-point lies at 1 while preserving the ratio between the durations implied by the sliders.  Figure \ref{fig:normalization} illustrates this procedure for three hypothetical annotators annotating the same two events $e_1$ and $e_2$. Assuming that the duration classes for $e_1$ or $e_2$ do not differ across annotators, the relative chronology of the events is the same in each case. This preservation of relative chronology, over absolute slider position, is important because, for the purposes of determining temporal relation, the absolute positions that annotators give are meaningless, and we do not want our models to be forced to fit to such irrelevant information. 

\paragraph{Inter-annotator agreement}
\label{sec:iaa}
We measure inter-annotator agreement (IAA) for the temporal relation sliders by calculating the rank (Spearman) correlation between the normalized slider positions for each pair of annotators that annotated the same group of five predicate pairs in the development set.\footnote{Our protocol design also allows us to detect some bad annotations internal to the annotation itself, as opposed to comparing one annotator's annotation of an item to another. See Appendix \ref{sec:data_collect_rejections} for further details on our deployment of such annotation-internal validation techniques.} The development set is annotated by 724 annotators. Rank correlation is a useful measure because it tells us how much different annotators agree of the relative position of each slider. The average rank correlation between annotators was 0.665 (95\% CI=[0.661, 0.669]).

For the duration responses, we compute the absolute difference in duration rank between the duration responses for each pair of annotators that annotated the same group of five predicate pairs in the development set. On average, annotators disagree by 2.24 scale points (95\% CI=[2.21, 2.25]), though there is heavy positive skew ($\gamma_1$ = 1.16, 95\% CI=[1.15, 1.18])---evidenced by the fact that the modal rank difference is 1 (25.3\% of the response pairs), with rank difference 0 as the next most likely (24.6\%) and rank difference 2 as a distant third (15.4\%). 

\paragraph{Summary statistics}

\begin{figure}[t]
\includegraphics[width=\columnwidth]{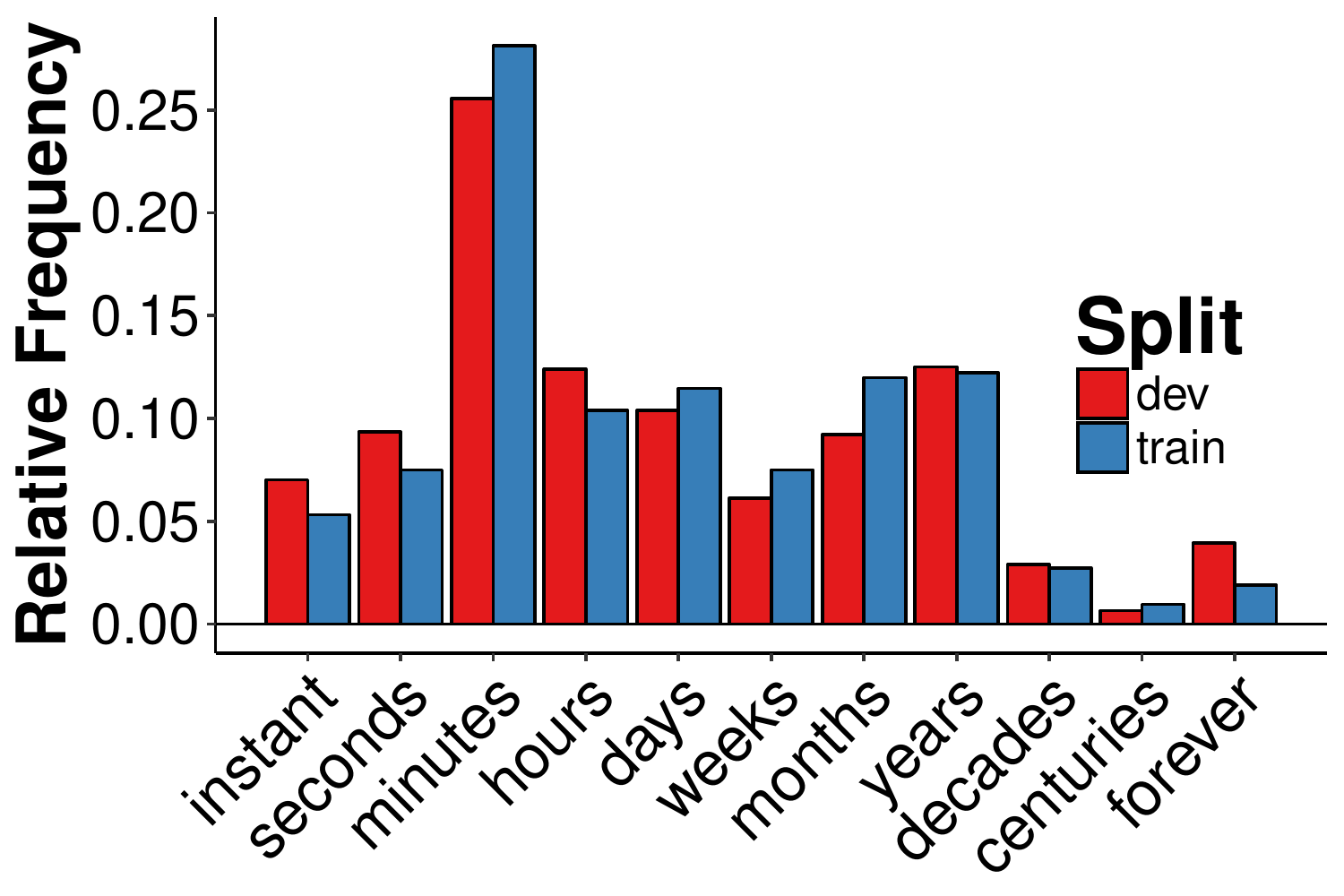}
\vspace{-8mm}
\caption{Distribution of event durations.}
\label{fig:duration_hist}
\vspace{-5mm}
\end{figure}

Figure \ref{fig:duration_hist} shows the distribution of duration responses in the training and development sets. There is a relatively high density of events lasting \textit{minutes}, with a relatively even distribution across durations of \textit{years} or less and few events lasting \textit{decades} or more.

The raw slider positions themselves are somewhat difficult to directly interpret.
To improve interpretability, we rotate the slider position space to construct four new dimensions: (i) \textsc{priority}, which is positive when $e_1$ starts and/or ends earlier than $e_2$ and negative otherwise; (ii) \textsc{containment},which is most positive when $e_1$ contains more of $e_2$; (iii) \textsc{equality}, which is largest when both $e_1$ and $e_2$ have the same temporal extents and smallest when they are most unequal; and (iv) \textsc{shift}, which moves the events forward or backward in time. We construct these dimensions by solving for $\mathbf{R}$ in 

\vspace{-7mm}

\[\mathbf{R}\begin{bmatrix*}[r]-1 & -1 &  1 &  1 \\
                           -1 &  1 &  1 & -1 \\
                           -1 &  1 & -1 &  1 \\
                            1 &  1 &  1 &  1\end{bmatrix*} = 2\mathbf{S} - 1\]

\vspace{-2mm}

\noindent where $\mathbf{S} \in [0, 1]^{N \times 4}$ contains the slider positions for our $N$ datapoints in the following order: beg($e_1$), end($e_1$), beg($e_2$), end($e_2$).

\begin{figure}[t]
\vspace{-3mm}
\includegraphics[width=\columnwidth]{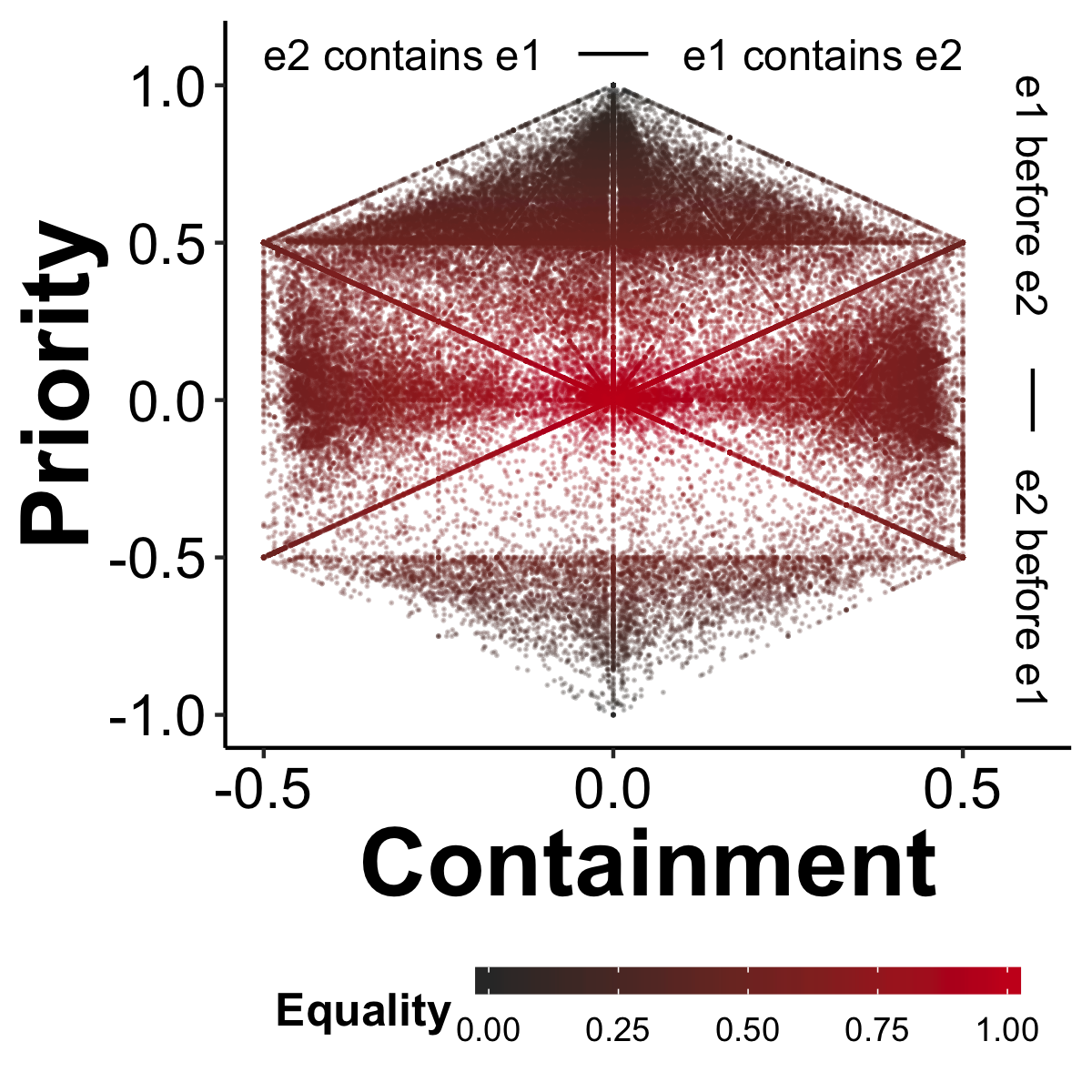}
\vspace{-10mm}
\caption{Distribution of event relations.}
\label{fig:relation_distr}
\vspace{-6mm}
\end{figure}

Figure \ref{fig:relation_distr} shows the embedding of the event pairs on the first three of these dimensions of $\mathbf{R}$. The triangular pattern near the top and bottom of the plot arises because strict priority---i.e. extreme positivity or negativity on the $y$-axis---precludes any temporal overlap between the two events, and as we move toward the center of the plot, different priority relations mix with different overlap relations---e.g. the upper-middle left corresponds to event pairs where most of $e_1$ comes toward the beginning of $e_2$, while the upper middle right of the plot corresponds to event pairs where most of $e_2$ comes toward the end of $e_1$.

\section{Model}
\label{sec:model}
For each pair of events referred to in a sentence, we aim to jointly predict the relative timelines of those events as well as their durations. We then use a separate model to induce document timelines from the relative timelines.

\paragraph{Relative timelines}

The relative timeline model consists of three components: an event model, a duration model, and a relation model. These components use multiple layers of \textit{dot product attention} \citep{luong2015effective} on top of an embedding $\Hb \in \reals^{N \times D}$ for a sentence $\ssb = [w_1, \ldots, w_N]$ tuned on the three $M$-dimensional contextual embeddings produced by ELMo \citep{Peters:2018} for that sentence, concatenated together.

\vspace{-6mm}
\[\Hb = \tanh\left(\text{ELMo}(\ssb)\Wb^\text{\sc tune} + \bbb^\text{\sc tune}\right)\]

\vspace{-2mm}

\noindent where $D$ is the dimension for the tuned embeddings, $\Wb^\text{\sc tune} \in \reals^{3M \times D}$, and $\bbb^\text{\sc tune} \in \reals^{N \times D}$.

\subparagraph{Event model}

We define the model's representation for the event referred to by predicate $k$ as $\gb_\text{pred$_k$} \in \reals^D$, where $D$ is the embedding size. We build this representation using a variant of dot-product attention, based on the predicate root.

\vspace{-5mm}

\begin{align*}
\ab^\text{\sc span}_\text{pred$_k$} &= \tanh\left(\Ab_\text{\sc pred}^\text{\sc span}\hb_\text{\textsc{root}(pred$_k$)} + \bbb_\text{\sc pred}^\text{\sc span}\right)\\
\bm{\alpha}_\text{pred$_k$} &= \mathrm{softmax}\left(\Hb_\text{\textsc{span}(pred$_k$)}\ab_\text{pred$_k$}^\text{\sc span}\right)\\
\gb_\text{pred$_k$} &= [\hb_\text{\textsc{root}(pred$_k$)}; \bm{\alpha}_\text{pred$_k$}\Hb_\text{\textsc{span}(pred$_k$)}]
\end{align*}


\noindent where $\Ab_\text{\sc pred}^\text{\sc span} \in \reals^{D \times D}, \bbb_\text{\sc pred}^\text{\sc span} \in \reals^D$; $\hb_\text{\textsc{root}(pred$_k$)}$ is the hidden representation of the $k^{th}$ predicate's root; and $\Hb_\text{\textsc{span}(pred$_k$)}$ is obtained by stacking the hidden representations of the entire predicate. 

\begin{figure}[t]
\includegraphics[width=\columnwidth]{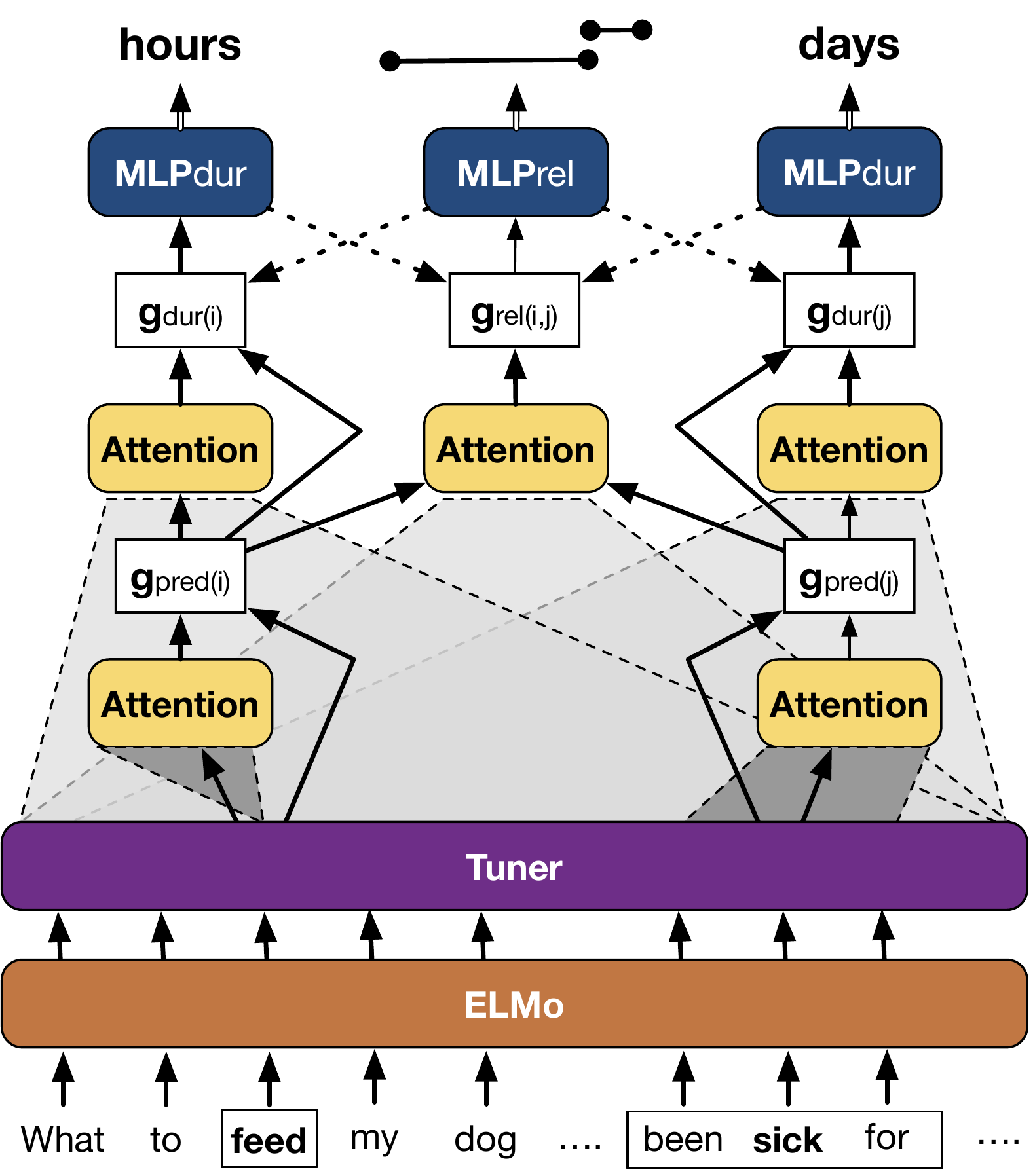}
\vspace{-6mm}
\caption{Network diagram for model. Dashed arrows are only included in some models.}
\vspace{-6mm}
\label{fig:modelfig}
\end{figure}

As an example, the predicate \textit{been \textbf{sick} for now} in Figure \ref{fig:protocolexample} has \textit{sick} as its root, and thus we would take the hidden representation for \textit{sick} as $\hb_\text{\textsc{root}(pred$_k$)}$. Similarly, $\Hb_\text{\textsc{span}(pred$_k$)}$ would be equal to taking the hidden-state representations of \textit{been sick for now} and stacking them together. Then, if the model learns that tense information is important, it may weight \textit{been} using attention.

\subparagraph{Duration model}
 
The temporal duration representation $\gb_\text{dur$_k$}$ for the event referred to by the $k^\text{th}$ predicate is defined similarly to the event representation, but instead of stacking the predicate's span, we stack the hidden representations of the entire sentence $\Hb$.

\vspace{-4mm}

\begin{align*}
\ab^\text{\sc sent}_\text{dur$_k$} &= \tanh\left(\Ab^\text{\sc sent}_\text{\sc dur}\gb_\text{pred$_k$} + \bbb^\text{\sc sent}_\text{\sc dur}\right)\\
\bm{\alpha}_\text{dur$_k$} 
&= \mathrm{softmax}(\Hb\ab^\text{\sc sent}_\text{dur$_k$})\\
\gb_\text{dur$_k$} &= [\gb_\text{pred$_k$} ; \bm{\alpha}_\text{dur$_k$}\Hb]
\end{align*}

\vspace{-1mm}

\noindent where $\Ab^\text{\sc sent}_\text{\sc dur} \in \reals^{D \times \text{size}(g_\text{pred$_k$})}$ and $\bbb^\text{\sc sent}_\text{\sc dur} \in \reals^D$.

We consider two models of the categorical durations: a softmax model and a binomial model. The main difference is that the binomial model enforces that the probabilities $\pb_\text{dur$_k$}$ over the 11 duration values be concave in the duration rank, whereas the softmax model has no such constraint. We employ a cross-entropy loss for both models. 

\vspace{-2mm}

\[\mathbb{L}_\text{dur}(d_k; \pb) = -\log p_{d_k}\]


\noindent In the softmax model, we pass the duration representation $\gb_\text{dur$_k$}$ for predicate $k$ through a multilayer perceptron (MLP) with a single hidden layer of ReLU activations, to yield probabilities $\pb_\text{dur$_k$}$ over the 11 durations. 

\vspace{-6mm}

\begin{align*}
\vb_\text{dur$_k$} &= \mathrm{ReLU}(\Wb^{(1)}_\text{\sc dur}\gb_\text{dur$_k$} + \bbb^{(1)}_\text{\sc dur})\\
\pb &= \mathrm{softmax}(\Wb^{(2)}_\text{\sc dur}\vb_\text{dur$_k$} + \bbb^{(2)}_\text{\sc dur}) \label{eq:dur_softmax}
\end{align*}

\vspace{-1mm}

\noindent In the binomial distribution model, we again pass the duration representation through a MLP with a single hidden layer of ReLU activations, but in this case, we yield only a single value $\pi_\text{dur$_k$}$. With $\vb_\text{dur$_k$}$ as defined above:

\vspace{-5mm}

\begin{align*}
\pi &= \sigma\left(\wb^{(2)}_\text{\sc dur}\vb_\text{dur$_k$} + \bbb^{(2)}_\text{\sc dur}\right)\\
p_c &= \binom{n}{c}\pi^n(1-\pi)^{(n-c)}
\end{align*} 

\vspace{-2mm}

\noindent where $c \in \{0,1,2,...,10\}$ represents the ranked durations -- instant (0), seconds (1), minutes (2), ..., centuries (9), forever (10) -- and $n$ is the maximum class rank (10). 

\subparagraph{Relation model}
To represent the temporal relation representation between the events referred to by the $i^\text{th}$ and $j^\text{th}$ predicate, we again use a similar attention mechanism. 

\vspace{-7mm}

\begin{align*}
    \ab^\text{\sc sent}_\text{rel$_{ij}$} &= \tanh\left(\Ab^\text{\sc sent}_\text{\sc rel}[\gb_\text{pred$_i$}; \gb_\text{pred$_j$}] + \bbb^\text{\sc sent}_\text{\sc rel}\right)\\
    \bm{\alpha}_\text{rel$_{ij}$} &= \mathrm{softmax}\left(\Hb\ab^\text{\sc sent}_\text{rel$_{ij}$}\right)\\
    \gb_\text{rel$_{ij}$} &= [\gb_\text{pred$_i$}; \gb_\text{pred$_j$}; \bm{\alpha}_\text{rel$_{ij}$}\Hb]
\end{align*}

\vspace{-2mm}

\noindent where $\Ab^\text{\sc sent}_\text{\sc rel} \in \reals^{D \times 2\text{size}(g_\text{pred$_k$})}$ and $\bbb^\text{\sc sent}_\text{\sc rel} \in \reals^D$.

The main idea behind our temporal model is to map events and states directly to a timeline, which we represent via a \textit{reference interval} $[0,1]$. For situation $k$, we aim to predict the beginning point $\text{b$_k$}$ and end-point $\text{e$_k$} \geq \text{b$_k$}$ of $k$.

We predict these values by passing $\gb_{rel_{ij}}$ through an MLP with one hidden layer of ReLU activations and four real-valued outputs $[\hat{\beta}_i, \hat{\delta}_i, \hat{\beta}_j, \hat{\delta}_j]$, representing the estimated relative beginning points ($\hat{\beta}_i, \hat{\beta}_j$) and durations ($\hat{\delta}_i, \hat{\delta}_j$) for events $i$ and $j$. We then calculate the predicted slider values $\hat{\ssb}_{ij} = [\hat{b}_i, \hat{e}_i, \hat{b}_j, \hat{e}_j]$

\vspace{-3mm}

\[\left[\hat{b}_k, \hat{e}_k\right] = \left[\sigma\left(\hat{\beta}_k\right), \sigma\left(\hat{\beta}_k + \left|\hat{\delta}_k\right|\right)\right]\]

\noindent The predicted values $\hat{\ssb}_{ij}$ are then normalized in the same fashion as the true slider values prior to being entered into the loss. We constrain this normalized $\hat{\ssb}_{ij}$ using four L1 losses.

\vspace{-5mm}

\begin{align*}
      \mathbb{L}_\text{rel}(\ssb_{ij}; \hat{\ssb}_{ij}) 
=& \left|(b_{i} - b_{j})-(\hat{b}_i-\hat{b}_{j})\right| +\\
& \left|(e_{i} - b_{j})-(\hat{e}_{i} -\hat{b}_{j})\right| +\\
& \left|(e_{j} - b_{i})-(\hat{e}_{j}  -\hat{b}_{i})\right| +\\
& \left|(e_{i} - e_{j})-(\hat{e}_{i} - \hat{e}_{j})\right|\\
  \end{align*}

\vspace{-7mm}

\noindent The final loss function is then $\mathbb{L} = \mathbb{L}_\text{dur} + 2\mathbb{L}_\text{rel}$.

\vspace{-2mm}

\subparagraph{Duration-relation connections}

We also experiment with four architectures wherein the duration and relation models are connected to each other in the Dur $\rightarrow$ Rel or Dur $\leftarrow$ Rel directions. 

In the Dur $\rightarrow$ Rel architectures, we modify $\gb_{rel_{ij}}$ in two ways:  (i) additionally concatenating the $i^{th}$ and $j^{th}$ predicate's duration probabilities from the binomial distribution model, and (ii) not using the relation representation model at all.

\vspace{-6mm}
\begin{align*}
\gb_\text{rel$_{ij}$} &= [\gb_\text{pred$_i$}; \gb_\text{pred$_j$}; \bm{\alpha}_\text{rel$_{ij}$}\Hb; \pb_i; \pb_j] \label{eq:grel_concat}\\
\gb_\text{rel$_{ij}$} &=  [\pb_i; \pb_j]
\end{align*}
\noindent In the Dur $\leftarrow$ Rel architectures, we use two modifications: (i) we modify $\gb_{dur_{k}}$ by concatenating the $\hat{b}_k$ and $\hat{e}_k$ from the relation model, and (ii) we do not use the duration representation model at all, instead use the predicted relative duration $\hat{e}_k-\hat{b}_k$ obtained from the relation model, passing it through the binomial distribution model.

\vspace{-8mm}
\begin{align*}
\gb_\text{dur$_k$} &= [\gb_\text{pred$_k$} ; \bm{\alpha}_\text{dur$_k$}\Hb; \hat{b}_k; \hat{e}_k]\\
\pi_\text{dur$_k$} &=  \hat{e}_k-\hat{b}_k 
\end{align*}

\vspace{-4mm}


\vspace{1mm}
\begin{table*}[t]
\begin{centering}
\begin{tabular}{lcc|ccc|ccc}
\toprule
\multicolumn{3}{c}{\textbf{Model}} & 
\multicolumn{3}{c}{\textbf{Duration}} & 
\multicolumn{3}{c}{\textbf{Relation}}\\
        \multicolumn{1}{c}{Duration} &
        \multicolumn{1}{c}{Relation} &
        \multicolumn{1}{c}{Connection} &
        \multicolumn{1}{c}{$\rho$} & 
        \multicolumn{1}{c}{rank diff.} &
        \multicolumn{1}{c}{R1} &
        \multicolumn{1}{c}{Absolute $\rho$} &
        \multicolumn{1}{c}{Relative $\rho$}  &
        \multicolumn{1}{c}{R1}\\
\midrule
softmax  & \checkmark & - & 32.63 & 1.86 & \phantom{-}8.59 & 77.91 & 68.00  & \phantom{-}2.82\\
\rowcolor{blue!10}binomial & \checkmark & - & 37.75 & \textbf{1.75} & \phantom{-}13.73 & 77.87 & 67.68 & \phantom{-}2.35\\
- & \checkmark & Dur $\leftarrow$ Rel & 22.65 & 3.08 & -51.68& 71.65 & 66.59 & -6.09\\
binomial & - & Dur $\rightarrow$ Rel & 36.52 & 1.76 & \phantom{-}13.17 & 77.58 & 66.36 & \phantom{-}0.85  \\
binomial & \checkmark & Dur $\rightarrow$ Rel & \textbf{38.38} & \textbf{1.75} & \phantom{-}\textbf{13.85} & 77.82 & 67.73 & \phantom{-}2.58  \\
binomial & \checkmark & Dur $\leftarrow$ Rel & 38.12 & 1.75 & \phantom{-}13.68 & \textbf{78.12} & \textbf{68.22} & \phantom{-}\textbf{2.96}  \\

\bottomrule
\end{tabular}

\caption{Results on test data based on different model representations; $\rho$ denotes the Spearman-correlation coefficient; rank-diff is the duration rank difference. The model highlighted in blue performs best on durations and is also close to the top performing model for relations on the development set. The numbers highlighted in \textbf{bold} are the best-performing numbers on the test data in the respective columns.}
\label{tab:experiments}
\vspace{-5mm}
\end{centering}
\end{table*}

\vspace{-3mm}
\paragraph{Document timelines}

We induce the hidden document timelines for the documents in the UDS-T development set using relative timelines from (i) actual pairwise slider annotations; or (ii) slider values predicted by the best performing model on UDS-T development set. To do this, we assume a hidden timeline $\Tb \in \reals_+^{n_d \times 2}$, where $n_d$ is the total number of predicates in that document, the two dimensions represent the beginning point and the duration of the predicates. We connect these latent timelines to the relative timelines, by anchoring the beginning points of all predicates such that there is always a predicate with 0 as the beginning point in a document and defining auxiliar variables $\bm{\tau}_{ij}$ and $\hat{\ssb}_{ij}$ for each events $i$ and $j$.

\vspace{-9mm}

\begin{align*}
    \bm{\tau}_{ij} &= [t_{i1}, t_{i1}+t_{i2}, t_{j1}, t_{j1}+t_{j2}]\\
    \hat{\ssb}_{ij} &= \frac{\bm{\tau}_{ij}-\min(\bm{\tau}_{ij})}{\max(\bm{\tau}_{ij}-\min(\bm{\tau}_{ij}))}\\
\end{align*}

\vspace{-7mm}

\noindent We learn $\Tb$ for each document under the relation loss $\mathbb{L}_\text{rel}(\ssb_{ij}, \hat{\ssb}_{ij})$. We further constrain $\Tb$ to predict the categorical durations using the binomial distribution model on the durations $t_{k2}$ implied by $\Tb$, assuming $\pi_k = \sigma(c\log(t_{k2}))$.

\section{Experiments}
\label{sec:experiments}

We implement all models in \texttt{pytorch 1.0}. For all experiments, we use mini-batch gradient descent with batch-size 64 to train the embedding tuner (reducing ELMo to a dimension of 256), attention, and MLP parameters. Both the relation and duration MLP have a single hidden layer with 128 nodes and a dropout probability of 0.5 (see Appendix \ref{sec:model_config} for further details).

To predict TimeML relations in TempEval3~\citep[TE3;][Task C-relation only]{uzzaman2013semeval}  and TimeBank-Dense~\citep[TD;][]{cassidy2014annotation}, we use a transfer learning approach. We first use the best-performing model on the UDS-T development set to obtain the relation representation ($\gb_{rel_{ij}}$) for each pair of annotated event-event relations in TE3 and TD (see Appendix \ref{sec:preprocess_timeml} for preprocessing details). We then use this vector as input features to a SVM classifier with a Gaussian kernel to train on
the training sets of these datasets using the feature vector obtained from our model.\footnote{For training on TE3, we use TimeBank~\citep[TB;][]{pustejovsky2003timebank} + AQUAINT~\citep[AQ;][]{graffaquaint} datasets provided in the TE3 workshop~\citep{uzzaman2013semeval}. For training on TD, we use TD-train and TD-dev.}


Following recent work using continuous labels in event factuality prediction ~\cite{lee2015event, stanovsky2017integrating, rudinger2018neural, white2018lexicosyntactic} and genericity prediction ~\cite{govindarajan_gener_2019}, we report three metrics for the duration prediction: Spearman correlation ($\rho$), mean rank difference (\textit{rank diff}), and proportion rank difference explained (R1). We report three metrics for the relation prediction: Spearman correlation between the normalized values of actual beginning and end points and the predicted ones (\textit{absolute} $\rho$), the Spearman correlation between the actual and predicted values in $\mathbb{L}_\text{rel}$ (\textit{relative} $\rho$), and the proportion of MAE explained (R1).


\[\text{R1} = 1-\frac{\text{MAE}_\text{model}}{\text{MAE}_\text{baseline}}\]


\noindent where $\text{MAE}_\text{baseline}$ is always guessing the median. 

\section{Results}
\label{sec:results}
Table \ref{tab:experiments} shows the results of different model architectures on the UDS-T test set, and Table \ref{tab:otherresults} shows the results of our transfer-learning approach on test set of TimeBank-Dense (TD-test). 

\paragraph{UDS-T results}

Most of our models are able to predict the relative position of the beginning and ending of events very well (high relation $\rho$) and the relative duration of events somewhat well (relatively low duration $\rho$), but they have a lot more trouble predicting relation exactly and relatively less trouble predicting duration exactly.

\subparagraph{Duration model} 

The binomial distribution model outperforms the softmax model for duration prediction by a large margin, though it has basically no effect on the accuracy of the relation model, with the binomial and softmax models performing comparably. This suggests that enforcing concavity in duration rank on the duration probabilities helps the model better predict durations. 

\begin{table*}[!t]
\centering
\begin{tabular}{ll}
    \begin{tabular}{lrrrr}
        \multicolumn{4}{c}{\textbf{Duration}}\\
        \toprule
        Word & 
        Attention &
        Rank & 
        Freq
        \\
        \midrule
        soldiers & 0.911 & 1.28 &  69\\
        \textbf{months} & 0.844 & 1.38 &  264\\
         Nothing & 0.777 & 5.07 &  114\\
         \textbf{minutes} & 0.768 & 1.33 &  81\\
         astronauts & 0.756 & 1.37 &  81\\
         \textbf{hour} & 0.749 & 1.41 & 84\\
         Palestinians & 0.735 & 1.72 &  288\\
         \textbf{month} & 0.721 & 2.03 &  186\\
         cartoonists & 0.714 & 1.35 &  63\\
         \textbf{years} & 0.708 & 1.94 &  588\\
         \textbf{days} & 0.635 & 1.39 &  84\\
         thoughts & 0.592 & 2.90 &  60\\
         us & 0.557 & 2.09 &  483\\
         \textbf{week} & 0.531 & 2.23 & 558\\
         advocates & 0.517 & 2.30 &  105\\
        \bottomrule
    \end{tabular}
    &
    \begin{tabular}{lrrrr}
    \multicolumn{4}{c}{\textbf{Relation}}\\
        \toprule
        Word & 
        Attention &
        Rank & 
        Freq
        \\
        \midrule
        \textbf{occupied} & 0.685 & 1.33 &  54\\
        massive & 0.522 & 2.71 &  66\\
        social & 0.510 & 1.68 &  57\\
        general & 0.410 & 3.52 &  168\\
        few & 0.394 & 3.07 &  474\\
        mathematical & 0.393 & 7.66 & 132\\
         \textbf{are} & 0.387 & 3.47 &  4415\\
         \textbf{comes} & 0.339 & 2.39 &  51\\
         \textbf{or} & 0.326 & 3.50 &  3137\\
         \textbf{and} & 0.307 & 4.86 &  17615\\
         emerge & 0.305 & 2.67 &  54\\
         \textbf{filed} & 0.303 & 7.14 &  66\\
         \textbf{s} & 0.298 & 4.03 &  1152\\
         \textbf{were} & 0.282 & 3.49 &  1308\\
         \textbf{gets} & 0.239 & 7.36 &  228\\
        \bottomrule
    \end{tabular}
\end{tabular}

\caption{Mean attention weight, mean attention rank, and frequency for 15 words in the development set with the highest mean duration-attention (left) and relation-attention (right) weights. For duration, the words highlighted in bold directly correspond to some duration class. For relation, the words in bold are either conjunctions or words containing tense information.}
\label{tab:attention_words}
\vspace{-5mm}
\end{table*}

\subparagraph{Connections} 

Connecting the duration and relation model does not improve performance in general. In fact, when the durations are directly predicted from the temporal relation model---i.e. without using the duration representation model---the model's performance drops by a large margin, with the Spearman correlation down by roughly 15 percentage points. This indicates that constraining the relations model to predict the durations is not enough and that the duration representation is needed to predict durations well. On the other hand, predicting temporal relations directly from the duration probability distribution---i.e. without using the relation representation model---results in a similar score as that of the top-performing model. This indicates that the duration representation is able to capture most of the relation characteristics of the sentence. Using both duration representation and relation representation separately (model highlighted in blue) results in the best performance overall on the UDS-T development set.

\paragraph{TimeBank-Dense and TempEval3}

Table \ref{tab:otherresults} reports F1-micro scores on the test set of TimeBank-Dense compared with some other systems as reported by \citet{cheng2017classifying}. We report these scores only on Event-Event (E-E) relations as our system captures only those. We also compute the standard temporal awareness F1 score on the test set of TempEval-3 (TE3-PT) considering only E-E relations and achieve a score of 0.498.\footnote{We do not report the \textit{temporal awareness} scores (F1) of other systems on TE3-PT, since they report their metrics on all relations, including timex-timex, and event-timex relations, and thus they are not directly comparable. For TD, only those systems are reported that report F1-micro scores.} Our system beats the TD F1-micro scores of all other systems reported in Table \ref{tab:otherresults}. As a reference, the top performing system on TE3-PT \citep{ning2017structured} reports an F1 score of 0.672 over all relations, but is not directly comparable to our system as we only evaluate on event-event relations.  These results indicate that our model is able to achieve competitive performance on other standard temporal classification problems. 

\begin{table}[h]
\small
\centering
\vspace{-2mm}
\begin{tabular}{llccc}
\toprule
\multicolumn{1}{c}{\textbf{Systems}} & 
\multicolumn{1}{c}{\textbf{Evaluation Data}} &
\multicolumn{1}{c}{\textbf{F1}} \\
    {} & {} &
    \multicolumn{1}{c}{\textbf{(E-E)}} \\
\midrule
CAEVO &  TD-test & 0.494   \\
CATENA & TD-test &  0.519  \\
\citet{cheng2017classifying} & TD-test & 0.529 \\
\textbf{This work} &  TD-test & \textbf{0.566}\\

\bottomrule
\end{tabular}






\vspace{-2mm}
\caption{F1-micro scores of event-event relations in TD-test based on our transfer learning experiment.}

\label{tab:otherresults}
\vspace{-5mm}
\end{table}

\section{Model Analysis and Timelines}
\label{sec:analysis}
We investigate two aspects of the best-performing model on the development set (highlighted in Table \ref{tab:experiments}): (i) what our duration and relation representations attend to; and (ii) how well document timelines constructed from the model's predictions match those constructed from the annotations. (See Appendix \ref{sec:appanalysis} for further analyses.)

\paragraph{Attention}

The advantage of using an attention mechanism is that we can often interpret what linguistic information the model is using by analyzing the attention weights. We extract these attention weights for both the duration representation and the relation representation from our best model on the development set. 

\subparagraph{Duration}

We find that words that denote some time period---e.g. \textit{month(s)}, \textit{minutes}, \textit{hour}, \textit{years}, \textit{days}, \textit{week}---are among the words with highest mean attention weight in the duration model, with seven of the top 15 words directly denoting one of the duration classes (Table \ref{tab:attention_words}). This is exactly what one might expect this model to rely heavily on, since time expressions are likely highly informative for making predictions about duration. It also may suggest that we do not need to directly encode relations between event-referring and time-referring expressions in our framework---as do annotation standards like TimeML---since our models may discover them.

The remainder of the top words in the duration model are plurals or mass nouns (\textit{soldiers, thoughts} etc.). This may suggest that the plurality of a predicate's arguments is an indicator of the likely duration of the event referred to by that predicate. To investigate this possibility, we compute a multinomial regression predicting the attention weights $\bm{\alpha}_s$ for each sentence $s$ from the $K$ morphological features of each word in that sentence $\Fb_s \in \{0, 1\}^{\text{length}(s) \times K}$, which are extracted from the UD-EWT features column and binarized. To do this, we optimize coefficients $\cb$ in $\arg_\cb\min \sum_s D\left(\bm{\alpha}_s\;\|\; \text{softmax}\left(\Fb_s\cb\right)\right)$, where $D$ is the KL divergence. We find that the five most strongly weighted positive features in $\cb$ are all features of nouns---\textsc{number}=\textit{plur}, \textsc{case}=\textit{acc}, \textsc{prontype}=\textit{prs}, \textsc{number}=\textit{sing}, \textsc{gender}=\textit{masc}---suggesting that good portion of duration information can be gleaned from the arguments of a predicate. This may be because nominal information can be useful in determining whether the clause is about particular events or generic events~\cite{govindarajan_gener_2019}.  

\vspace{-2mm}
\subparagraph{Relation}

A majority of the words with highest mean attention weight in the relation model are either coordinators---such as \textit{or} and \textit{and}---or bearers of tense information---i.e. lexical verbs and auxiliaries. The first makes sense because, in context, coordinators can carry information about temporal sequencing~\citep[see][i.a.]{wilson1998pragmatics}. The second makes sense in that information about the tense of predicates being compared likely helps the model determine relative ordering of the events they refer to.  

Similar to duration attention analysis, for relation attention, 
we find that the five most strongly weighted positive features in $\cb$ are all features of verbs or auxiliaries---\textsc{person}=1,
\textsc{person}=3,
\textsc{tense}=\textit{pres},
\textsc{tense}=\textit{past},
\textsc{mood}=\textit{ind}---suggesting that a majority of the information relevant to relation can be gleaned from the tense-bearing units in a clause.

\begin{figure}[t]
\includegraphics[width=\columnwidth]{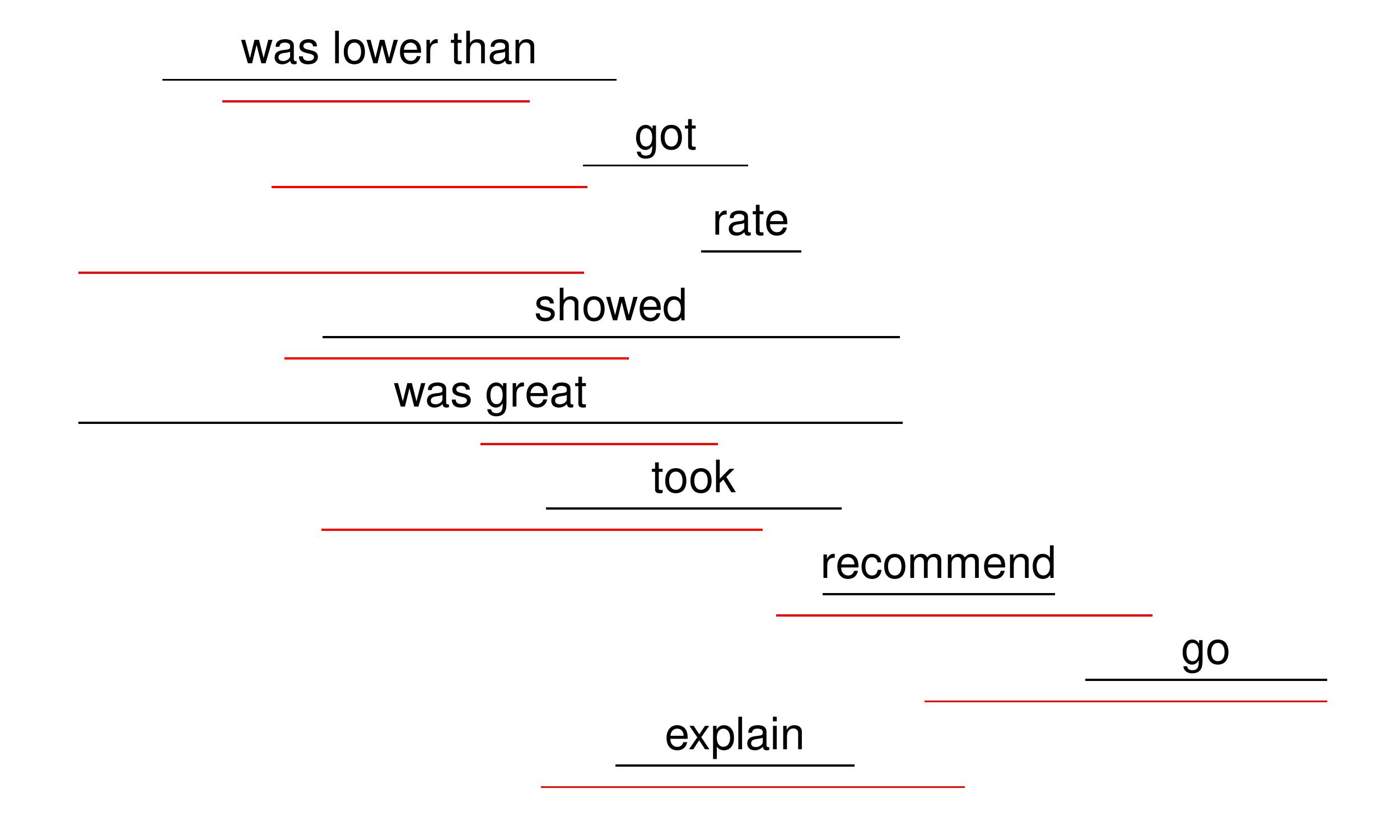}
\vspace{-10mm}
\caption{Learned timeline for the following document based on actual (black) and predicted (red) annotations:
\textit{``A+. I would rate Fran pcs an A + because the price was lower than everyone else , i got my computer back the next day , and the professionalism he showed was great .
He took the time to explain things to me about my computer , i would recommend you go to him. David"}}
\label{fig:timeline}
\vspace{-6mm}
\end{figure}

\paragraph{Document timelines}

We apply the document timeline model described in \S\ref{sec:model} to both the annotations on the development set and the best-performing model's predictions to obtain timelines for all documents in the development set. Figure \ref{fig:timeline} shows an example, comparing the two resulting document timelines.


For these two timelines, we compare the induced beginning points and durations, obtaining a mean Spearman correlation of 0.28 for beginning points and -0.097 for durations. This suggests that the model agrees to some extent with the annotations about the beginning points of events in most documents but is struggling to find the correct duration spans. One possible reason for poor prediction of durations could be the lack of a direct source of duration information. The model currently tries to identify the duration based only on the slider values, which leads to poor performance as already seen in one of the Dur $\leftarrow$ Rel model.

\section{Conclusion}
\label{sec:conclusion}
We presented a novel semantic framework for modeling fine-grained temporal relations and event durations that maps pairs of events to real-valued scales for the purpose of constructing document-level event timelines. We used this framework to construct the largest temporal relations dataset to date -- UDS-T -- covering the entirety of the UD-EWT.  We used this dataset to train models for jointly predicting fine-grained temporal relations and event durations, reporting strong results on our data and showing the efficacy of a transfer-learning approach for predicting standard, categorical TimeML relations. 

\section*{Acknowledgments}
\label{sec:acknowledgments}
We are grateful to the FACTS.lab at the University of Rochester as well as three anonymous reviewers for useful comments on this work. This research was supported by the University of Rochester, JHU HLTCOE, and DARPA AIDA.  The U.S. Government is authorized to reproduce and distribute reprints for Governmental purposes. The views and conclusions contained in this publication are those of the authors and should not be interpreted as representing official policies or endorsements of DARPA or the U.S. Government.

\bibliography{acl2019}
\bibliographystyle{acl_natbib}

\pagebreak
\clearpage

\appendix
\section{Data Collection}
\label{sec:data_collect_appendix}
\begin{figure}[h]
\includegraphics[width=\columnwidth]{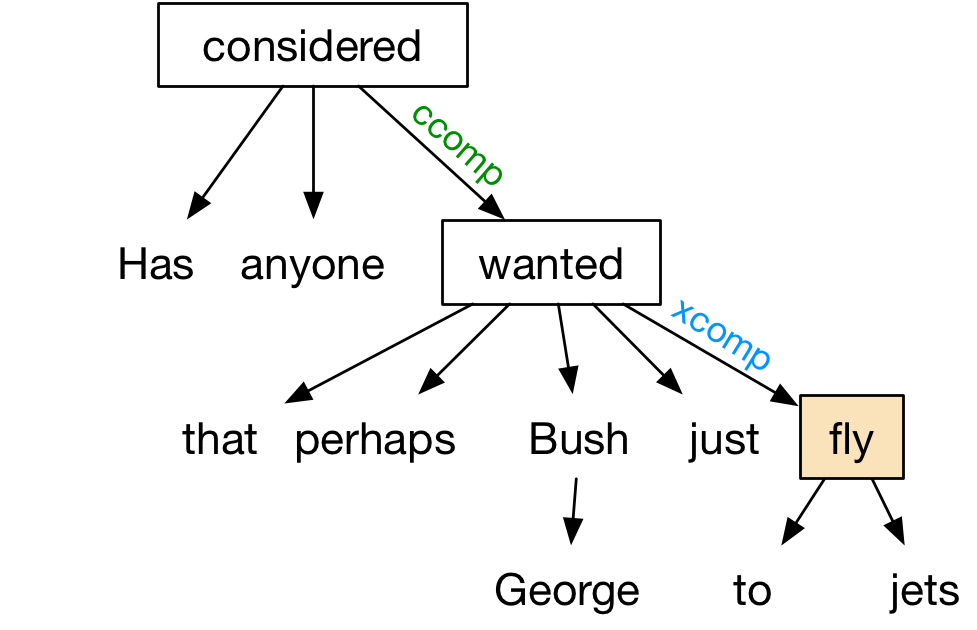}
\vspace{-9mm}
\caption{Our heuristic finds \textit{fly} as (the root of) the pivot predicate in \textit{Has anyone considered that perhaps George Bush just wanted to fly jets?}}
\vspace{-6mm}
\label{fig:pivotpred}
\end{figure}

We concatenate two adjacent sentences to form a combined sentence which allows us to capture inter-sentential temporal relations. Considering all possible pairs of events in the combined sentence results into an exploding number of event-event comparisons. Therefore, to reduce the total number of comparisons, we find the \textit{pivot-predicate} of the antecedent of the combined sentence as follows - find the root predicate of the antecedent and if it governs a \texttt{CCOMP}, \texttt{CSUBJ}, or \texttt{XCOMP}, follow that dependency to the next predicate until a predicate is found that doesn't govern a \texttt{CCOMP}, \texttt{CSUBJ}, or \texttt{XCOMP}. We then take all pairs of the antecedent predicates and pair every predicate of the consequent only with the \textit{pivot-predicate}. This results into $\binom{N}{2} + M$ predicates instead of $\binom{N+M}{2}$ per sentence, where N and M are the number of predicates in the antecedent and consequent respectively.  This heuristic allows us to find a predicate that loosely denotes the topic being talked about in the sentence. Figure \ref{fig:pivotpred} shows an example of finding the pivot predicate.

\section{Rejecting Annotations}
\label{sec:data_collect_rejections}
\begin{figure}[t]
\includegraphics[width=\columnwidth]{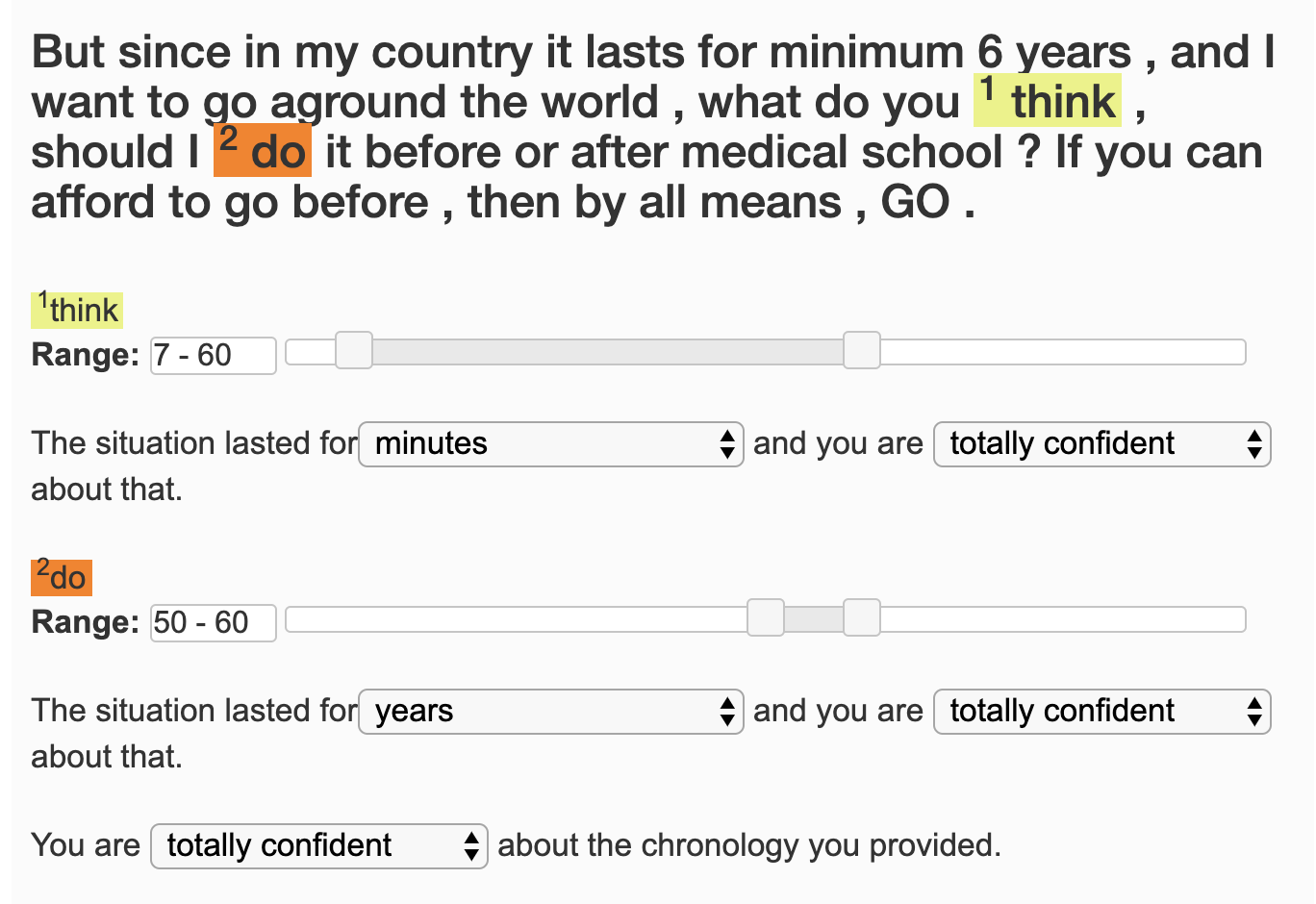}
\vspace{-9mm}
\caption{An example illustrating an inconsistency between the annotated slider positions and the durations}
\vspace{-6mm}
\label{fig:inconsistency}
\end{figure}
We design multiple checks to detect potentially bad annotations during our data collection. A single assignment contains 5 annotations (predicate-pairs). Once an annotation is flagged by any of these checks, we may accept or reject the assignment based on our subjective opinion about the particular case. Annotations are flagged based on the following conditions:
\subsection{Time completion}
Our pilot studies indicate a median time of roughly 4 minutes to complete a single assignment (5 annotations). We automatically reject any assignment which is completed under a minute as we believe that it is not plausible to finish the assignment within a minute. We find that such annotations mostly had default values annotated. 

\subsection{Same slider values}
If all the beginning points and end-points in an assignment have the same values, we automatically reject those assignments.

\subsection{Same duration values}
Sometimes we encounter cases where all duration values in an assignment are annotated to have the same value. This scenario , although unlikely, could genuinely be an instance of correct annotation. Hence we manually check for these cases and reject only if the annotations look dubious in nature based on our subjective opinion. 

\subsection{Inconsistency between the slider positions and durations}
Our protocol design allows us to detect potentially bad annotations by detecting inconsistency between the slider positions (beginning and end-points) and the duration values of events in an annotated sentence. The annotator in Figure \ref{fig:inconsistency} assigns slider values for $e_1$ (\textit{think}) as [7,60] i.e. a time-span of 53 and assigns its duration as \textit{minutes}. But at the same time, the slider values for $e_2$ (\textit{do}) are annotated as [50,60] i.e. a time-span of 10, even though its duration is assigned as \textit{years}. This is an inconsistency as $e_2$ has a smaller time-span denoted by the sliders but has the longer duration as denoted by \textit{years}. We reject assignments where more than 60\% of annotations have this inconsistency.

\section{Inter-annotator agreement}
\label{sec:interanno_appendix}

Annotators were asked to approximate the relative duration of the two events that they were annotating using the distance between the sliders. This means that an annotation is coherent insofar as the ratio of distances between the slider responses for each event matches the ratio of the categorical duration responses. We rejected annotations wherein there was gross mismatch between the categorical responses and the slider responses --- i.e. one event is annotated as having a longer duration but is given a shorter slider response --- but because this does not guarantee that the exact ratios are preserved, we assess that here using a canonical correlation analysis (CCA; \citealt{hotelling_relations_1936}) between the categorical duration responses and the slider responses. 

\begin{figure}[htb!]
\vspace{-2mm}
\includegraphics[width=\columnwidth]{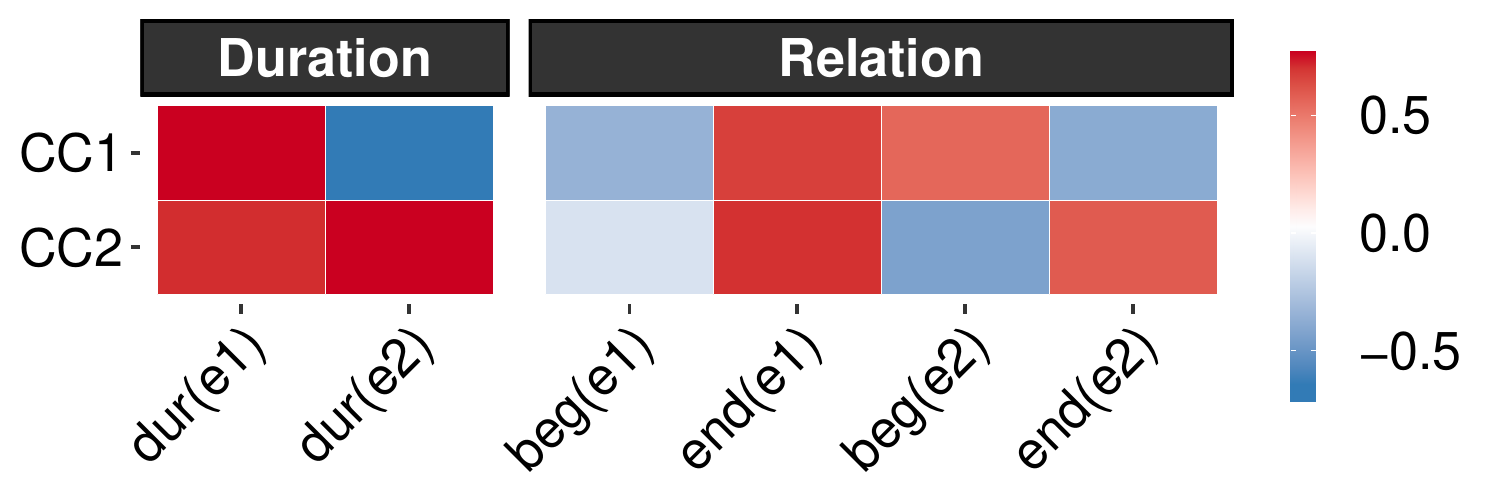}
\vspace{-6mm}
\caption{Scores from canonical correlation analysis comparing categorical duration annotations and slider relation annotations.}
\vspace{-4mm}
\label{fig:ccascores}
\end{figure}

Figure \ref{fig:ccascores} shows the CCA scores. We find that the first canonical correlation, which captures the ratios between unequal events, is 0.765; and the second, which captures the ratios between roughly unequal events, is 0.427. This preservation of the ratios is quite impressive in light of the fact that our slider scales are bounded; though we hoped for at least a non-linear relationship between the categorical durations and the slider distances, we did not expect such a strong linear relationship.

\section{Confidence Ratings}
\label{sec:model_config}
Annotators use the confidence scale in different ways. Some always respond with \textit{totally confident} whereas others use all five options. To cater to these differences, we normalize the confidence ratings for each event-pair using a standard ordinal scale normalization technique known as ridit scoring. In ridit scoring ordinal labels are mapped to (0, 1) using the empirical cumulative distribution function of the ratings given by each annotator. Ridit scoring re-weights the importance of a scale label based on the frequency of its usage.

We weight both $\mathbb{L}_\text{dur}$, and $\mathbb{L}_\text{rel}$ by the ridit-scored confidence ratings of event durations and event relations, respectively. 

\section{Processing TempEval3 and TimeBank-Dense}
\label{sec:preprocess_timeml}
Since we require spans of predicates for our model, we pre-process TB+AQ and TD by removing all xml tags from the sentences and then we pass it through Stanford \texttt{CoreNLP 3.9.2} ~\citep{manning2014stanford} to get the corresponding conllu format. Roots and spans of predicates are then extracted using PredPatt. To train the SVM classifier, we use \texttt{sklearn 0.20.0};~\citealt{pedregosa_scikit-learn:_2011}. We run a hyperparameter grid-search over 4-fold CV with C: (0.1, 1, 10), and gamma: (0.001, 0.01, 0.1, 1). The best performance on cross-validation (C=1 and gamma=0.001) is then evaluated on the test set of TE3 i.e. TE3-Platinum (TE3-PT),  and TD-test. For our purposes, the \textit{identity} and \textit{simultaneous} relations in TB+AQ are equivalent when comparing event-event relations. Hence, they are collapsed into one single relation. 

\section{Further analysis}
\label{sec:appanalysis}
We rotate the predicted slider positions in the relation space defined in \S\ref{sec:datacollection} and compare it with the rotated space of actual slider positions. We see a Spearman correlation of 0.19 for \textsc{priority}, 0.23 for \textsc{containment}, and 0.17 for \textsc{equality}. This suggests that our model is best able to capture \textsc{containment} relations and slightly less good at capturing \textsc{priority} and \textsc{equality} relations, though all the numbers are quite low compared to the \textit{absolute} $\rho$ and \textit{relative} $\rho$ metrics reported in Table \ref{tab:experiments}. This may be indicative of the fact that our models do somewhat poorly on predicting more fine-grained aspects of an event relation, and in the future it may be useful to jointly train against the more interpretable \textsc{priority}, \textsc{containment}, and \textsc{equality} measures instead of or in conjunction with the slider values.




\end{document}